\documentclass{poster09}
\pdfoutput=1

\usepackage[utf8]{inputenc}

\usepackage{listings}
\usepackage{hyperref}
\hypersetup{pdfauthor={Ivo Danihelka},pdftitle={Optimistic Simulated Exploration as an Incentive for Real Exploration}}

\lstnewenvironment{smallexample}[1]{%
\lstset{language={python}, showstringspaces=false, basicstyle=\footnotesize\ttfamily, frame=single, inputencoding=utf8, captionpos=b, #1}}
{
\vspace{1em}
}

\begin{document}


%
\title{Optimistic Simulated Exploration as an Incentive for Real Exploration}
%
\headtitle{I. DANIHELKA, OPTIMISTIC SIMULATED EXPLORATION AS AN INCENTIVE FOR REAL EXPLORATION}

%
\author{Ivo DANIHELKA\affiliationmark{1}}
%
\affiliation{%
\affiliationmark{1}Dept. of Cybernetics, Czech Technical University, Karlovo náměstí 13, 121 35 Prague 2, Czech Republic}
  \email{ivo.danihelka@agents.felk.cvut.cz}

\maketitle


\begin{abstract}
Many reinforcement learning exploration techniques are overly optimistic and
try to explore every state.
Such exploration is impossible in environments
with the unlimited number of states.
I propose to use simulated exploration with an optimistic model
to discover promising paths for real exploration.
This reduces the needs for the real exploration.
\end{abstract}



\section{Introduction}
In reinforcement learning\cite{rl_book} an agent collects
rewards in an environment. The environment is not known in advance.
The agent has to explore it to learn where to go.

A reward could be received when taking an action in a state.
The agent aims to maximize her long-term reward in the environment.
She should not miss any state with an important reward or a shorter path to it.

There are many existing exploration techniques\cite{OIM} that are optimistic
in the face of uncertainty. Their optimism assumes that a greater reward will be obtained when taking an unknown action.

The problem is how to do exploration in environments with the unlimited
number of states. In these environments, it is not possible to try
every action in every possible state.

I study a new way to do exploration in environments with the unlimited number of states. I use simulated exploration as an incentive for real exploration.
The simulated exploration proposes promising paths to explore.
I describe how to use this kind of exploration in section \ref{Idea}.

My experiments in section \ref{Experiments}
demonstrate how simulated exploration reduced
the needed amount of real exploration.
I also discuss when it is possible.

Many works touched related problems. They inspired me
and I discuss them in section \ref{RelatedWork}.

\subsection{An Example of the Problem}
\begin{figure}[ht!]
\begin{center}
\resizebox{65mm}{!}{\includegraphics{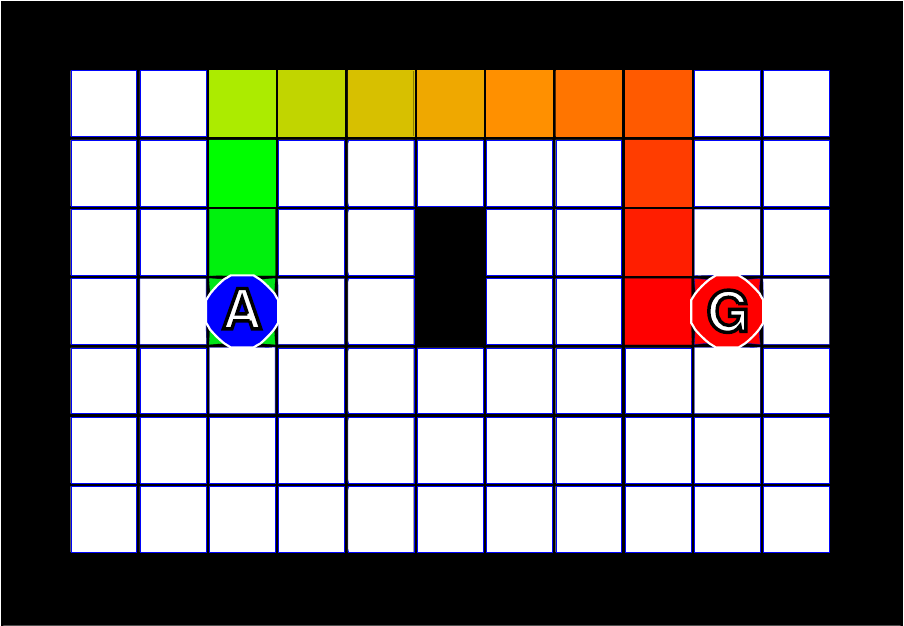}}
\caption{A grid world environment with agent \textit{A}, goal state \textit{G} and a suboptimal path from the start to the goal.}
\label{img_example}
\end{center}
\end{figure}

There is an example of a reinforcement learning task in Figure \ref{img_example}. The environment is a grid world. The agent has to find a path from the start to the goal. This environment gives a reward just when the goal is reached.
The task is then repeated.

A state is represented as a tile on the grid
and the agent could choose from four different actions:
\textit{North}, \textit{East}, \textit{South}, \textit{West}. But the agent does not know their meanings.

The agent does not know that the environment is a 2D grid world
with Manhattan distance metric.
She sees it just as a graph and has yet to explore its structure.

A \textit{value} of a state represents the summed expected
reward that the agent expects to see if she continues to follow her policy form that state.
Q-learning\cite{rl_survey} is typically used to learn the values of state-action transitions.

In an environment with the limited number of states,
We could use exploration that is optimistic about rewards from unknown actions.
This exploration would try to explore every action in every state
to see if there is a greater reward.
Such wide exploration is impossible if the environment is without boundaries.

\section{The Idea}
\label{Idea}
The idea is to use simulated exploration to discover promising paths for real
exploration. This reduces the needs for real exploration.
An approximate model of the environment is used to simulate the exploration.

\subsection{Approximate Models}
An approximate model is an approximation of the real environment.
It predicts how the environment will behave when executing an action
in a state. The model predicts the next state of the environment
and what reward will be obtained for the action.

An approximate model could be optimistic or pessimistic.
An optimistic model assume a greater reward or better transitions than
it is possible in the real environment.
For example, an optimistic model could assume no barriers on the path.

It is better to use optimistic models than the pessimistic ones\cite{rl_book}.
Using an optimistic model will lead to discovery of a more accurate model
of the environment or to discovery of better paths.
A pessimistic model would assume that no better path exists
and would miss it.

The optimistic model should be as accurate as possible
to prevent too many mistakes and corrections.
An overly optimistic model would assume that every state transition will
lead to a state with the highest reward.
A model could aim to be more accurate at the risk to become pessimistic
in some states. That is a risk we accept when we don't want
to explore all states in unlimited state space.

For example, a model aiming to be more accurate
could assume that already tried actions
will have similar effects in new states. This does not need to be true
when a wall is hit, but that would be corrected if the model is optimistic.
An example of one such model is given in section \ref{Experiments}.

These less optimistic models may not need to work as an incentive
for exploration on their own. A greedy agent sees no reason
to try new actions when they don't lead directly to states with a higher value.
A simulated exploration is needed to discover the promising paths.

\subsection{Simulated Exploration}
When given an approximate model,
the simulated exploration will try to find promising paths under that model.
It tries different actions in the model
and explores where they lead. The exploration is not real,
it is simulated without taking such actions in the real environment.

To work as an incentive for real exploration,
values of environment states continue to be updated during the simulation
as if it were a real experience. When an unexplored state leads to a state with a high value,
the value of the unexplored state is also increased.

The simulation is done as any other planning method.
It is executed to do a few simulated explorations
and then it is interleaved by real acting.
The following code shows a body of a typical agent.
The simulated exploration would be inside the \texttt{self.planner.plan()}
method, possibly with some other planning.

\begin{smallexample}{caption={Body of a typical agent expressed in Python.}}
def agent_step(self, s, r=None):
  """Does a single step and
  returns a selected action.
  The agent sees the current state `s' and
  the reward `r' obtained for the last action.
  """
  if r is not None:
    self.learner.learn(
        self.last_s, self.last_a, r, s)

  if rl.is_terminal(s):
    return None

  self.planner.plan(s)

  a = self.policy.select_action(s)
  self.last_s = s
  self.last_a = a
  return self.last_a
\end{smallexample}

The simulated exploration could discover a promising path
that leads to an existing or predicted reward.
The promising path would be visible
for the agent as a set of states that got a high value.
Some of these states could still be unexplored in the real environment.

I will describe two ways how to spread the simulated exploration:
\textit{trajectory sampling} and \textit{prioritized sweeping}.
But any method that will simulate experience within a given approximate model
could be used.

\subsubsection{Trajectory Sampling}
The simulated exploration could follow a trajectory generated
by an exploration policy. The trajectory could start from any state.
For example, it could start from the agent's real current state.
That restricts the simulated state space to the states near the agent.

Code \ref{code_sampling} shows an example of simulated exploration along a trajectory.
The maximal depth of the sampled trajectory is limited to limit the amount
of computation done inside one planning step.

\begin{smallexample}{caption={Simulating exploration by trajectory sampling.},label=code_sampling}
def plan_along_trajectory(self, s):
  """Simulates exploration of a trajectory
  from the given state.
  """
  path = []
  while len(path) <= self.max_depth:
    if rl.is_terminal(s):
      break

    a = self.exploration_policy.select_action(s)
    next_s_probs, r = self.model.predict(s, a)
    if len(next_s_probs) == 0:
      break

    path.append((s, a))
    next_s = choose_probable(next_s_probs)
    if next_s == s:
      break
    s = next_s

  for s, a in reversed(path):
    self.updater.estimate_q(s, a)
\end{smallexample}

The simulated exploration policy
could be completely different from the policy used for acting
in the real environment. The simulated exploration policy
is just used to sample states from the state space.
The approximate model is used to estimate values of the sampled states.

\subsubsection{Prioritized Sweeping}
Prioritized sweeping\cite{improved_sweeping} schedules updates of \mbox{state-action} values
when one of their children changed its value. A priority queue is typically used for that. The amount of change in a child serves as a priority to process its parents.

Prioritized sweeping could be used to simulate exploration on its own
or in combination with other methods.
When used on its own, it is needed to have an approximate model that is
able to return parents of a state. Related unexplored states
have to be also returned as possible parents.

The depth of sweeping of unexplored states is limited by setting a
minimal considered priority. The states where the amount
of change is below this threshold are not swept. This will work
unless the approximate model repeatedly predicts a reward
in new unexplored states.

When the model does not give information about parents,
it is still possible to use prioritized sweeping in combination
with another simulated exploration. It is just needed to correct the remembered parents
when the optimistic model is corrected. Otherwise a wrong parent
would continue to be updated by optimistic values.

Remembered parents could be easily corrected when the used model
produces a distribution of all possible next states. These next states
are all possible children of a parent and its old children could be discarded.

\section{Experiments}
\label{Experiments}
The experiments test how simulated exploration
reduces the amount of needed real exploration. It is assumed that
a suitable optimistic model of the environment could be used.

\subsection{The Task}
\begin{figure}[ht!]
\begin{center}
\resizebox{65mm}{!}{\includegraphics{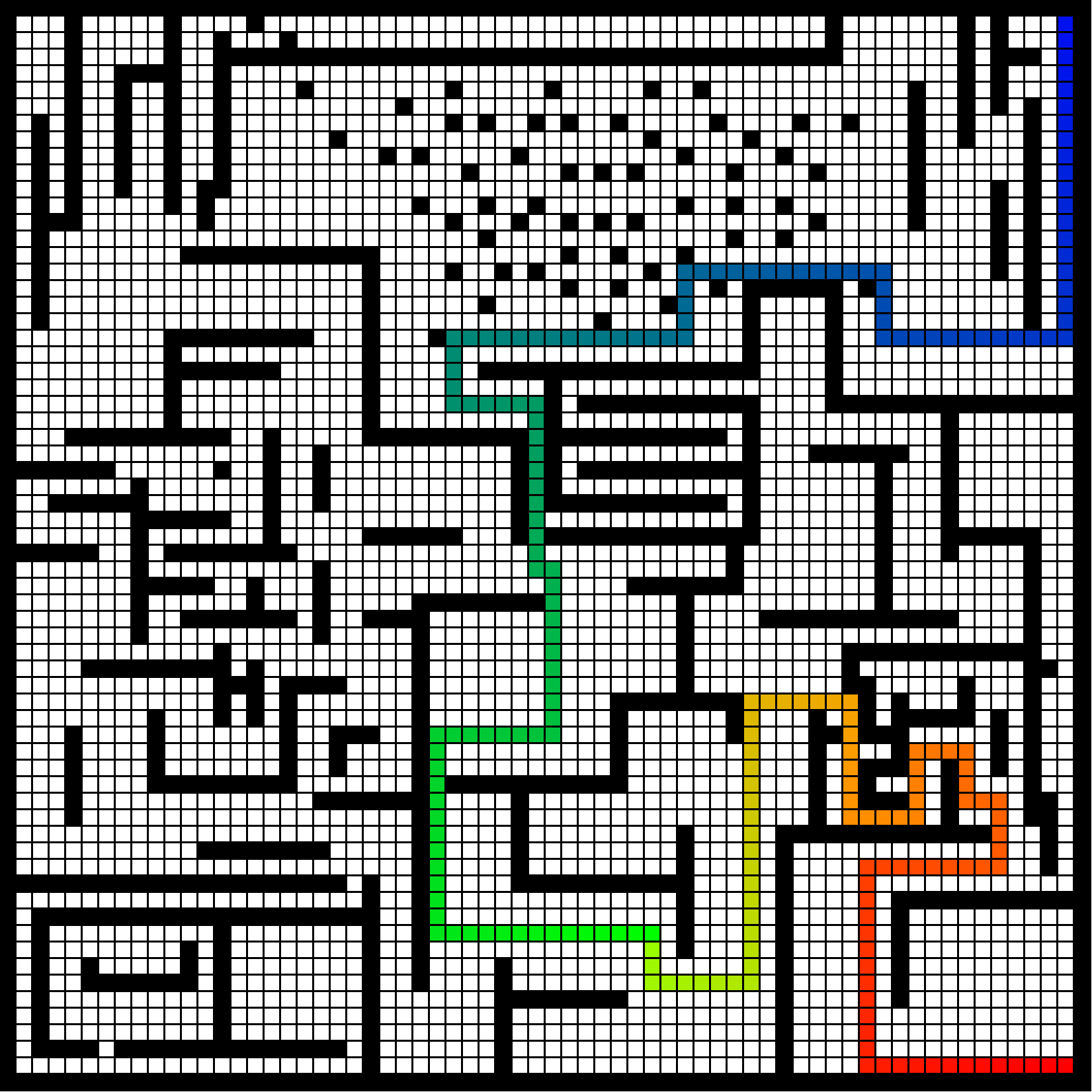}}
\caption{A grid world used for experiments. It contains an initial suboptimal path from the start to the goal.}
\label{img_grid_maze}
\end{center}
\end{figure}

In my experiments, I used the 3277-state grid world mentioned inside the \textit{Reinforcement Learning: A Survey}\cite{rl_survey}. Figure \ref{img_grid_maze} shows the used environment. The environment is fully deterministic.

I specified an initial suboptimal path inside the grid world. The suboptimal path gives a hint where the reward is in the unlimited state space. This allows to use the simulated exploration without optimistic assumptions about the reward.

The start position of the agent is fixed to the top-right corner of the grid world. The aim to reduce the amount of exploration
is most relevant when the start position is fixed.
In that case, the state space used for acting is smaller than the whole
available state space.

\subsection{The Used Model}
The used approximate model was composed as a combination two models:
model based on observations and a model based on recent action effects.

The \textbf{model based on observations} is a classical model that records frequencies of different outcomes for every state-action transition.
The frequencies are used to estimate the probabilities of the outcomes.
This model does not predict anything about yet unexplored states.

The \textbf{model based on recent action effects} records the last
seen effects of each action. It then predicts that a seen action
will have the same effects in a new state. It also tries to be optimistic
and ignores when an action does not have any effects on the state.
That could happen when a wall is hit. So it does not remember any wall.
The following code shows a code of such a model.

\begin{smallexample}{caption={A model that tries to predict state changes.}}
class RecentEffectBasedModel:
  def __init__(self, actions, initial_r):
    self.actions = actions
    self.initial_r = initial_r
    self.increments = [None] * len(actions)

  def predict(self, s, a):
    """Predicts probabilities of next states
    and the returned reward.
    """
    a_index = self.actions.index(a)
    increment = self.increments[a_index]
    if increment is None:
      return [], 0

    return [(s + increment, 1.0)], self.initial_r

  def learn(self, s, a, r, next_s):
    """Remembers state changes from experience.
    It remembers the last seen increment
    for the given action.
    """
    if rl.is_terminal(next_s):
      return

    # Walls are not learned by this model.
    # It is kept optimistic for the grid world.
    increment = next_s - s
    if increment != 0:
      a_index = self.actions.index(a)
      self.increments[a_index] = increment
\end{smallexample}

The above shown code assumes that a state is represented
by a single numeric variable. But an analogous approach
could be used when the state is represented by a vector of multiple variables\cite{structure}.

The used initial suboptimal path should use every action at least once.
That allows the model based on action effects to predict unexplored states in all directions. An alternative would be to do an additional
small amount of exploration at the beginning of the task.

The two used models provide different predictions about state transitions
and rewards. The model based on observations predicts just already
seen transitions and rewards. The model based on recent action effects
tries to predict future state transitions and assumes no extra reward from them.
The prediction based on recent action effects is wrong when a wall is hit,
so it is considered to be less accurate.

These two models are combined together to compose a final approximate model.
The final model uses the most accurate prediction available.
It asks the models for a prediction starting from the most accurate model.
When the asked model knows nothing about the given transition,
a less accurate model is asked. The code of the combined model follows.

\begin{smallexample}{caption={A model that combines multiple approximate models.}}
class CombinedModel:
  def __init__(self, models):
    """Accepts a given list of models
    that have decreasing accuracy.
    """
    self.models = models

  def predict(self, s, a):
    """Returns the most accurate
    available prediction
    for the given transition.
    """
    for model in self.models:
      next_s_probs, r = model.predict(s, a)
      if len(next_s_probs) > 0:
        return next_s_probs, r
    return [], 0

  def learn(self, s, a, r, next_s):
    for model in self.models:
      model.learn(s, a, r, next_s)
\end{smallexample}

\subsection{Used Algorithms}
I compare exploration with optimistic initial values
with two simulated explorations: trajectory sampling
and prioritized sweeping of unexplored states.

The exploration with \textbf{optimistic initial values} assumes
value 1.0 inside every state-action transition. This strategy
is incentive enough to find an optimal policy in a deterministic environment.

The initial state-action values for the other exploration algorithms
were set to $10^{-32}$. Such a low value does not serve as an incentive for exploration, but it remains possible to decrease it. It is important to be able to decrease the value of tried state-action transitions. It prevents the agent to get stuck.

The simulated \textbf{trajectory sampling} was tested with three different
maximal depths: 3, 6 and 12. The used simulated exploration policy selects a random action for every step. The sampled trajectory always starts from the current agent's state. The simulated exploration tries 10 trajectories per each real step.

The prioritized \textbf{sweeping of unexplored} states uses parents
supplied by the approximate model of the environment.
Our approximate model consists of multiple models, so
it has to combine the parents predicted by the different models.
I do this by letting the least accurate model predict some
parents. The more accurate models are then used to prune the list of possible
parents. The second least accurate model then predicts some other parents.
The algorithm is depicted by Code \ref{code_get_parents}.

\begin{smallexample}{caption={A method to return combined parents from multiple models. The used models have decreasing accuracy.},label=code_get_parents}
def get_parents(self, s):
  """Returns possible parents of the given state.
  """
  parents = set()
  for i, model in reversed(list(
        enumerate(self.models))):
    subparents = model.get_parents(s)
    for parent in subparents:
      if is_possible_transition(parent, s,
            self.models[:i]):
        parents.add(parent)

  return parents
\end{smallexample}

All mentioned explorations were aided with prioritized sweeping of explored states.
I implemented the improved prioritized sweeping algorithm of Wiering and Schmidhuber\cite{improved_sweeping}.

\subsection{Results}
I measured the amount of exploration done under the different exploration algorithms. The exploration is finished when the shortest path is found and used for subsequent episodes. The initial path was 217 steps long. The shortest path was 203 steps long and all used algorithms were able to find it.

The numbers of explored states and state-actions are given in table \ref{tab_results}.

\begin{table}[h]
\begin{center}
{\renewcommand{\arraystretch}{1.4}
\tabcolsep=.2cm
\tablefont
\begin{tabular}{|l|r|r|r|}
\hline
&Steps
&States
&State-actions\\
\hline
Optimistic initial values
&19,673
&3,201
&12,797\\
\hline
Trajectory sampling (3)
&2,830
&523
&902\\
\hline
Trajectory sampling (6)
&4,967
&932
&1,864\\
\hline
Trajectory sampling (12)
&5,291
&1,695
&3,668\\
\hline
Sweeping of unexplored
&6,372
&1,580
&3,536\\
\hline
\end{tabular}}
\caption{The performance of different explorations.
The columns report the number of steps until finding the final policy
and the numbers of explored states and state-actions.
The trajectory sampling used maximal depths 3, 6 and 12.
}
\label{tab_results}
\end{center}
\end{table}

It is interesting to note the relation between
the number of explored states and the number of explored state-actions.
The exploration equipped with simulated exploration
tried on average about two actions in every visited state.
On the other hand, the exploration with optimistic initial values
tried all four actions in almost every visited state.

\subsection{Discussion}
All measured simulated explorations had lower amount of exploration
than the overly optimistic exploration with optimistic initial values.

The trajectory sampling required the lowest amount of exploration to find the shortest path. But it does not guarantee that it will find the shortest path. The maximal depth of the sampled trajectory limits the space where to search for promising paths. That limits the amount of exploration done,
but it also allows to miss an optimal path.

The prioritized sweeping of unexplored states performed very well.
It required to explore 3.6 times less \mbox{state-actions} than the exploration with optimistic initial values. And the sweeping of unexplored states guarantees to find
any possible promising path that exists under the used approximate model.

The lower amount of real exploration was possible
thanks to simulated exploration in an approximate optimistic model.
These methods are not suitable for environments where it is impossible
to learn an optimistic model of unexplored states of the environments.
These methods risk that they will miss the optimal path when a pessimistic model is used.

The used approximate model does not predict any unexplored reward.
It was needed to start with an initial suboptimal path to show some reward.
More sophisticated approximate models could also try to predict the reward.

\section{Related Work}
\label{RelatedWork}
The using of an observed model for planning was pioneered by the Dyna architecture by Sutton\cite{dyna}. The Dyna planning continues to update seen transitions with changes
in the estimated state-action value. Its extensions provide prioritized sweeping\cite{improved_sweeping} and a usage with linear function approximation to represent the environment\cite{dyna_fn_approximation}.

The idea to use optimistic models came from the book \textit{Reinforcement Learning: An Introduction}\cite{rl_book}. It discusses how optimistic models do not miss a promising path. It also proposes to use trajectory sampling on large tasks.

Many works touched the problem of exploration in large spaces.
Smart and Kaelbling\cite{practical_rl} reduced the amount of exploration
by using initial knowledge. They supplied the agent with example
trajectories. During these trajectories, the agent was driven by a human operator or by a piece of code.
The example trajectories do not needed to be optimal. They just have to
give hints where the reward is.

Envelope methods\cite{envelope} limit the amount of planning by restricting the
state space of a known environment. They do not need to explore, because the environment is fully known, but they still needed to limit the number of states to consider.

Apprenticeship learning\cite{apprenticeship} aims to prevent destructive exploration. A teacher first demonstrates the task. That demonstration serves to learn an approximate model of environment. An optimal policy is learned off-line in the model and tested later in the real environment. The new experience serves to improve the model and the cycle continues. The apprenticeship learning is not intended to propose new promising paths, so unexplored states are not considered in the model.

An extension to the apprenticeship learning uses the learned approximate model
to find a policy improvement direction\cite{inaccurate_models}.
It could be viewed as a search for a promising direction
where to steer the policy parameters. The needed amount of steering
is then tested in the real environment.

\section{Conclusions}
I proposed a new way how to do exploration in unlimited state spaces.
It uses a simulated exploration and an optimistic model of the environment.

It remains to be discovered how to learn optimistic models
for a wide range of environments.
They should provide optimistic predictions about unexplored states,
but their optimism should be as low as possible.

My experiments show the possibility to reduce the real exploration
when given an optimistic model.

\section*{Acknowledgements}
I thank Jefferson Provost for releasing PLASTK (Python Learning Agent Software Toolkit). It allowed me to do pleasant experimenting with various environments.

I also wish to thank the community around the \mbox{rl-list}@googlegroups.com mailing list for provocative ideas and discussions.


\begin{authorcv}{Ivo DANIHELKA}
is a PhD student at the Department of Cybernetics at CTU in Prague.
Previously, he was a software engineer working on billing systems,
TV over the Internet, games, data mining and multiple web sites.
He still likes to read well written source code.
\end{authorcv}

\end{document}